\documentclass[lettersize,journal]{IEEEtran}
\usepackage{amsmath,amsfonts}
\usepackage{algorithmic}
\usepackage{algorithm}
\usepackage{array}
\usepackage{xfrac}
\usepackage[caption=false,font=normalsize,labelfont=sf,textfont=sf]{subfig}
\usepackage{textcomp}
\usepackage{stfloats}
\usepackage{url}
\usepackage{verbatim}
\usepackage{graphicx}
\usepackage{cite}
\usepackage{bbm}
\usepackage{multirow}
\usepackage{amsthm}
\usepackage{amssymb}
\usepackage{hyperref}
\usepackage{url}     
\usepackage{nicefrac}
\usepackage{xcolor}
\usepackage{backref}
\usepackage{xspace}
\usepackage{bm}
\usepackage{float}

\newtheorem{proposition}{Proposition}
\newtheorem{theorem}{Theorem}

\newtheorem{corollary}{Corollary}
\newtheorem{definition}{Definition}
\newtheorem{example}{Example}
\newtheorem{assumption}{Assumption}

\DeclareMathOperator{\Pos}{Pos}
\DeclareMathOperator{\CE}{CE}
\DeclareMathOperator{\ent}{H}
\DeclareMathOperator{\dist}{dist}

\begin{document}

\newcommand{\name}{\text{PEOPL}\xspace}

\title{\name: Characterizing Privately Encoded Open Datasets with Public Labels}

\author{%
  Homa~Esfahanizadeh, Adam~Yala, Rafael~G.~L. D’Oliveira, Andrea~J.~D.~Jaba, Victor~Quach, Ken~R.~Duffy, Tommi~S.~Jaakkola,  Vinod~Vaikuntanathan, Manya Ghobadi, Regina Barzilay, Muriel M\'{e}dard
  \thanks{H.~Esfahanizadeh, A.~Jaba, V.~Quach, T.~Jaakkola, V.~Vaikuntanathan, M.~Ghobadi, R.~Barzilay, and M~M\'{e}dard are with the Electrical Engineering and Computer Science Department, Massachusetts Institute of Technology (MIT), Cambridge, MA USA (emails: \{homaesf@, adjaba@, quach@, tommi@, vinodv@, ghobadi@csail., regina@csail., medard@\}mit.edu). A.~Yala is with University of California, Berkeley and University of California, San Francisco, CA USA (email: yala@berkeley.edu). R.~D’Oliveira is with School of Mathematical and Statistical Sciences, Clemson University, SC USA. K.~Duffy is with Maynooth University, Ireland (email: Ken.Duffy@mu.ie). Research supported in part by the National Science Foundation under grant no. CNS-2008624, and in part by Wellcome Trust Fellowship and MIT J-Clinic.}
}

\maketitle

\begin{abstract}
Allowing organizations to share their data for training of machine learning (ML) models without unintended information leakage is an open problem in practice. A promising technique for this still-open problem is to train models on the encoded data. Our approach, called Privately Encoded Open Datasets with Public Labels (\name), uses a certain class of randomly constructed transforms to encode sensitive data. Organizations publish their randomly encoded data and associated raw labels for ML training, where training is done without knowledge of the encoding realization. We investigate several important aspects of this problem: We introduce information-theoretic scores for privacy and utility, which quantify the average performance of an unfaithful user (e.g., adversary) and a faithful user (e.g., model developer) that have access to the published encoded data. We then theoretically characterize primitives in building families of encoding schemes that motivate the use of random deep neural networks. Empirically, we compare the performance of our randomized encoding scheme and a linear scheme to a suite of computational attacks, and we also show that our scheme achieves competitive prediction accuracy to raw-sample baselines. Moreover, we demonstrate that multiple institutions, using independent random encoders, can collaborate to train improved ML models.
\end{abstract}


\section{Introduction} 
For many applications, training Machine Learning (ML) models requires advanced and expensive computational resources and a large set of labeled data with sufficient diversity. An attractive approach is to enable co-operation across organizations by outsourcing the training data from several centers to the cloud where the model is trained. However, privacy concerns have complicated the construction of multi-institution cohorts and limited the utilization of external ML resources. For example, to protect patient privacy, regulations such as HIPPA~\cite{HIPAA} and GDPR~\cite{GDPR} prohibits sharing patients' identifiable information. Similarly, companies may be reluctant to share data that is part of their intellectual property and provides them a competitive edge. It is well-understood now that merely removing metadata, e.g., patient's name, is not enough for hiding sensitive information \cite{RocherNature2019}. Characterizing and mitigating this challenge is the main focus of this paper. We are interested in developing a computationally effective mechanism to encode sensitive date in order to facilitate outsourcing of training for ML. Inherent in our model is the notion that labels will be perforce \textit{public} but that we wish to reduce the information leakage beyond the release of labels.

\begin{figure*}
    \centering
    \begin{tabular}{cc}
    \includegraphics[width=0.47\textwidth]{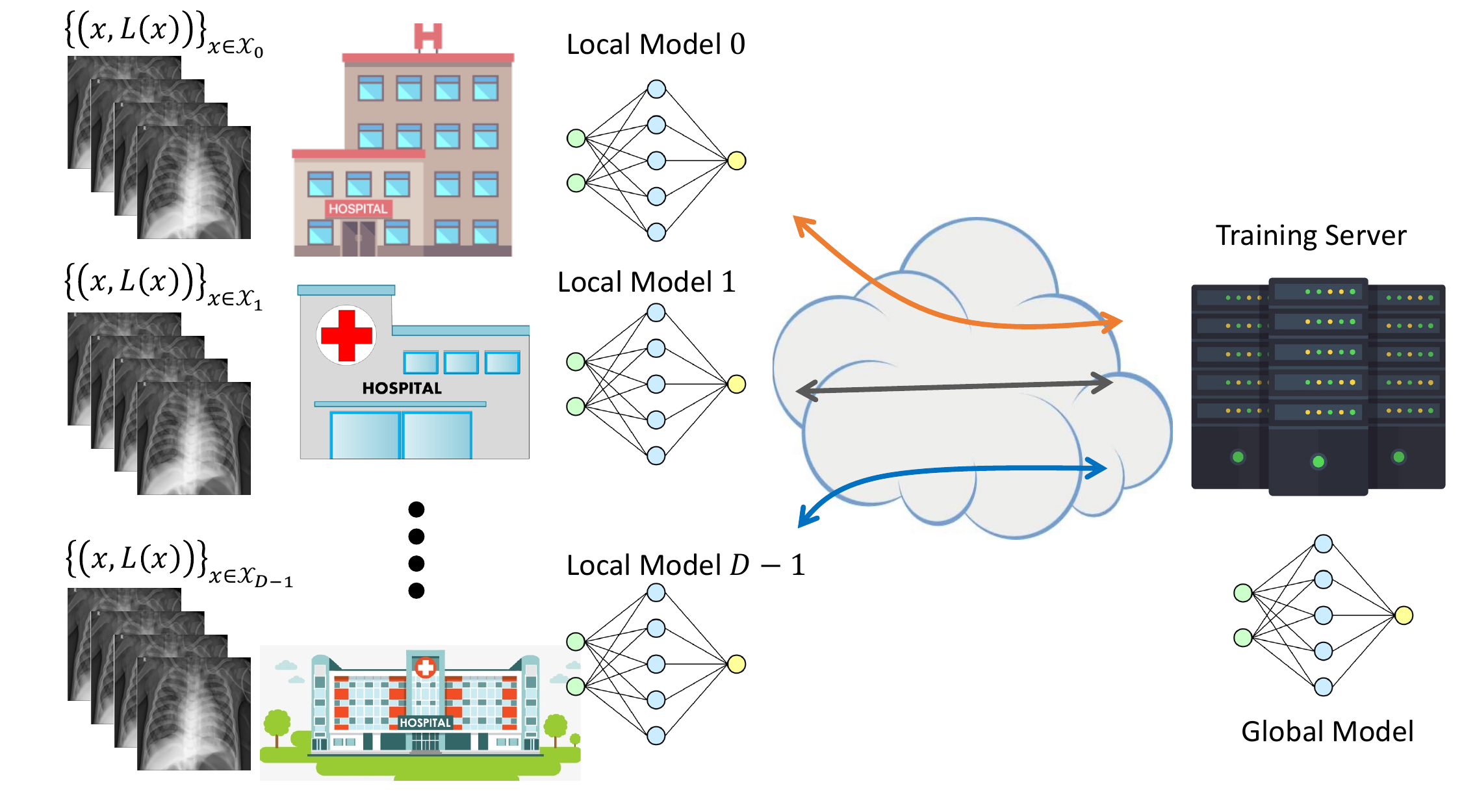}&
    \includegraphics[width=0.47\textwidth]{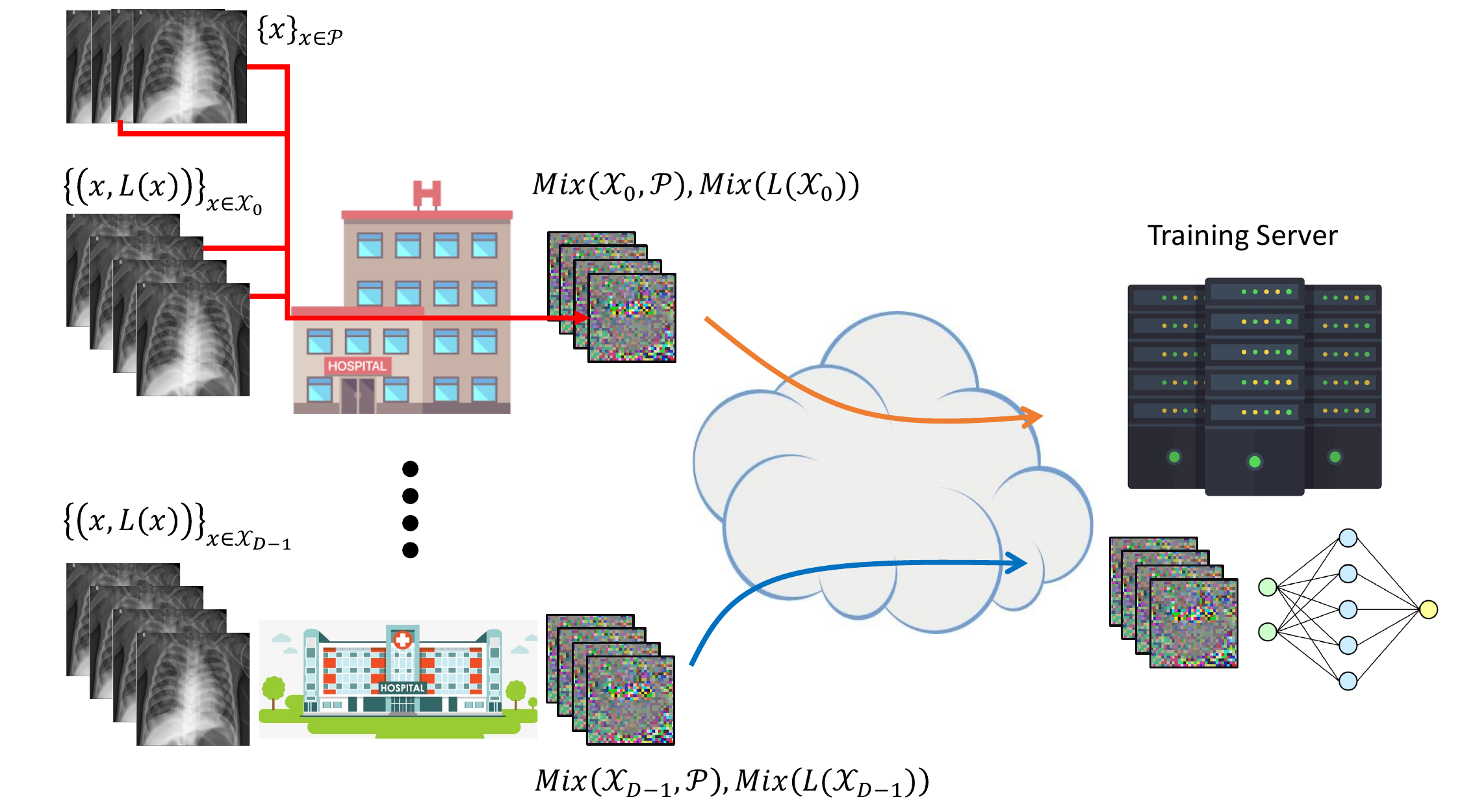}\\
    (a) Federated learning~\cite{mcmahan2017communication}&(b) Instahide~\cite{instahide} \\
    \multicolumn{2}{c}{\includegraphics[width=0.47\textwidth]{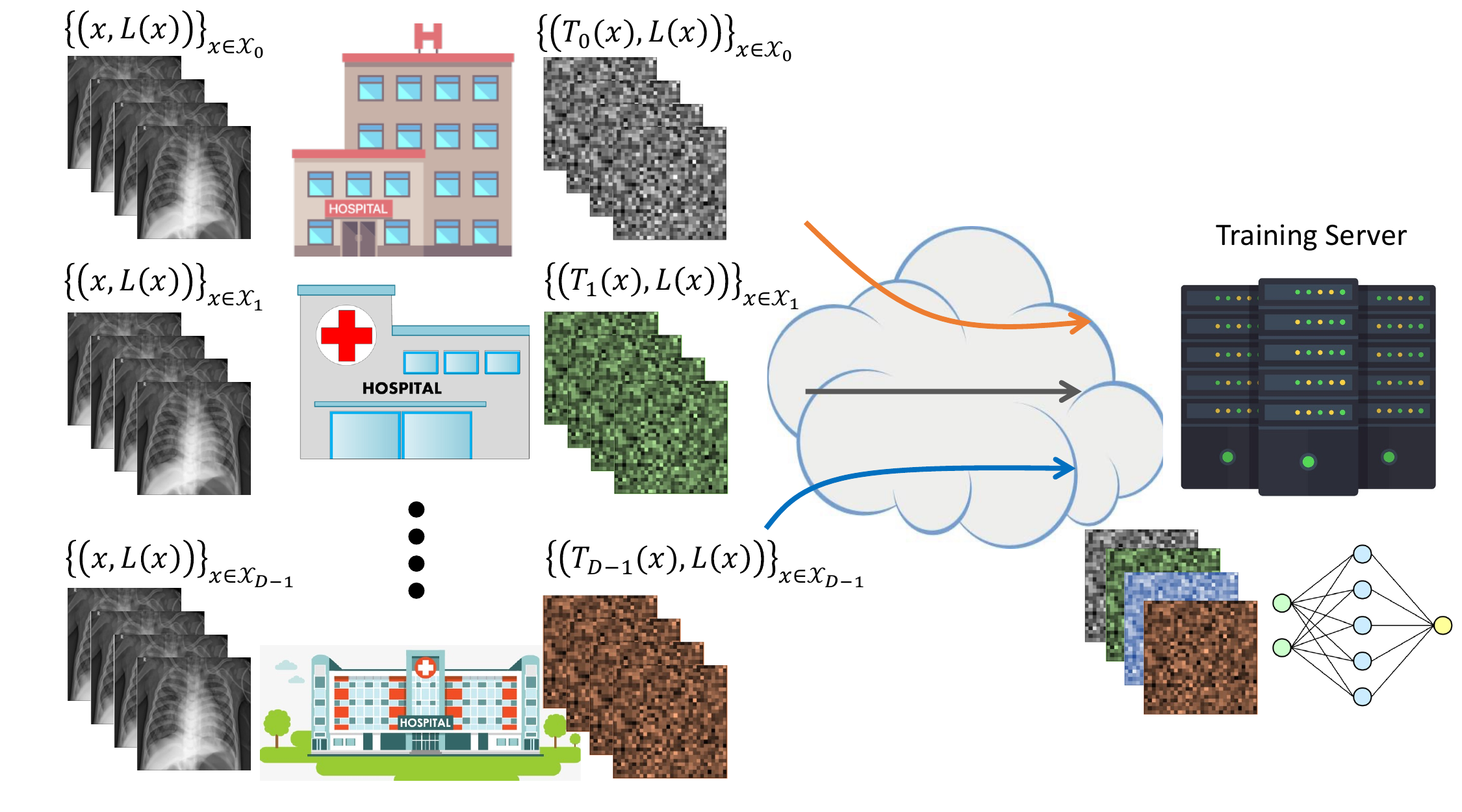}}\\
    \multicolumn{2}{c}{(c) This paper: \name}
    \end{tabular}
    \caption{Three approaches for mitigating privacy concerns in outsourced training and collaborative learning. (a) Federated learning \cite{mcmahan2017communication}: There exists a copy of global model at each data-owner. The local models are updated using data of their data-owner, and the model updates are exchanged with the server, rather than the sensitive data. (b) Instahide \cite{instahide}: The sensitive samples are linearly mixed with each other and with some public samples. The mixed samples and mixed labels are transferred to the server for training. (c) \name: Each sample is encoded by a nonlinear transform dedicated to the data-owner. The encoders of data-owners are sampled from a random distribution independently from each other, and they do not need to be shared with other data-owners or with the server. The encoded samples and raw labels are transferred to the server for training.\vspace{-0.5cm}}
    \label{fig:comp}
\end{figure*}

There are a number of solutions that have been developed for different notions of privacy. For instance, federated learning \cite{mcmahan2017communication} trains models in a distributed fashion, and in conjunction with adding noise during training \cite{Adnan2022}, it can obtain the theoretical notion of differential privacy \cite{dwork2014algorithmic}, see Fig.~\ref{fig:comp}~(a). However federated learning frameworks require a tight coordination across data-owners and model developers to jointly perform the training. Thereby, such approaches are not suitable for enabling data-owners to deposit their datasets publicly. As another instance, cryptographic methods such as secure multi-party computation, fully homomorphic encryption, and functional encryption \cite{Gentry09,BV11,BSW12,cho2018secure}  enable public sharing and offer extremely strong security guarantees by hiding everything about the data. However, these security guarantees come at the cost of extremely high computational and communication overheads for training today's advanced ML models \cite{cryptoeprint:2011/566,Cheetah}. Moreover, the cryptographic methods do not accommodate the  collaborative setting, i.e., training a single classifier using data of multiple data-owners, unless the data-owners trust each other and share the same key.

We argue that common ML training tasks do not require a strong level of security to hide everything about the data. An example is the training task of an ML model for diagnosing medical complications from chest x-ray images. The training task already implies the information that most images in the dataset contain human 24 ribs without looking into individual ones. Consequently, the notion of security adopted by the cryptographic methods is an overkill for the outsourced ML training task, where labels are public. Instead, we seek an efficient encoding scheme to protect the information that is not already implied by the general information about the training task and the samples' labels. 

Random encoders were recently considered in the literature to train models directly on encoded data~\cite{instahide,DAUnTLeSS,Syfer}. Instahide~\cite{instahide} used random linear mixing of images in the private dataset and some public dataset to generate the encoded samples and encoded labels. It demonstrated the feasibility of training models on the randomly transformed data and its potentials, see Fig.~\ref{fig:comp}~(b). However, as we show in this paper, the linearity of this scheme renders it vulnerable to adversarial distribution attacks. DauntLess~\cite{DAUnTLeSS} used random dense neural networks for encoding the samples and random mixing for encoding the labels. The authors proved perfect privacy is feasible when the encoding scheme is data-dependent, which is costly and not aligned with the philosophy of the outsourced training and collaborative learning. 

In this paper, we characterize randomized encoding schemes used by data-owners to publish their \textit{encoded} data, with associated \textit{uncoded} labels, from both privacy and utility perspectives. While developing practical schemes that guarantee that the encoded data cannot be used for any purpose other than the the designated training task (i.e., perfect schemes) remains an open challenge, our theoretical results offer primitives for improving scheme privacy. Building on these insights, we present \name, an encoding scheme based on the random selection of an encoder from a rich family of neural networks. We empirically demonstrate that \name obtains improved privacy over linear baselines while obtaining competitive accuracy to raw-sample baselines.  When applied in the multi-institutional setting, each site uses independent, uncoordinated random encoders to encode their data. With the help of label information, models trained in this setting can map these independently constructed encodings into a shared feature space, see Fig.~\ref{fig:comp}~(c). Our empirical results do not guarantee privacy against all possible attackers and \name should not be used in real-world sensitive settings today. However, these results do reflect improved privacy-utility trade-offs compared to common baselines and our analysis demonstrates a promising new direction for scheme design.

The organization of this paper are as follows: In Section~\ref{sec:prob-scores}, we formulate the problem of hiding sensitive data via a random transform. We propose the notions of privacy score and utility score, using the Shannon entropy \cite{Shannon48}, that quantify the performance of the probability distribution, according to which the random transform is chosen. In Section~\ref{sec:theory}, we propose how to improve the private key distribution to obtain a better privacy score. In particular, we show that function composition maintains or increases the privacy score. Furthermore, we study two types of attacks on the encoder that is chosen by a data-owner, given the available information to the adversary: The optimal attack which is based on the actual probability distribution, and a sub-optimal attack which is based on a mismatched distribution.

In Section~\ref{sec:method}, motivated by our theoretical results, we implement the family of possible encoding functions as random convolutional neural networks and random recurrent neural networks, for encoding the images and the texts, respectively. Our implemented schemes are designed for the cases where sensitive data does not have overlap with the relevant publicly-available datasets. We hypothesized that these schemes offer improved privacy over linear approaches. In Section~\ref{sec:exp}, we empirically compare our method to linear baselines on two chest x-ray image datasets MIMIC-CXR \cite{johnson2019mimic}, CheXpert \cite{irvin2019chexpert}, and one text dataset SMS spam collection \cite{spamdata}. Section~\ref{sec:Discussion} and \ref{sec:conclusion} are dedicated to discussion and conclusions, respectively.

\section{Problem Setting and Performance Scores}\label{sec:prob-scores}

We first introduce necessary notations and definitions. We denote a set of elements with Calligraphic letters, e.g., $\mathcal{X}$. A transformation (encoder) is denoted with $T:\mathcal{X} \rightarrow \mathcal{Z}$. The cardinality (the number of elements) of a set and the factorial function are denoted by $|.|$ and $(.)!$, respectively. For notation purposes, we occasionally impose a total order $\preceq$ on $\mathcal{X}$ to represent it by a vector $(x_1,\ldots,x_{|\mathcal{X}|})$ such that $x_i \preceq x_j$ if $i\leq j$. We then can represent an encoder by a vector with size $|\mathcal{X}|$ whose $i$-th element is $T(x_i)$.  Without loss of generality, we assume all logarithms are in base $2$ in this paper. The mathematical derivations used rely on both stochastic (random) and deterministic variables. All stochastic variables are denoted with bold font, e.g., $\textbf{x}$, while deterministic variables are denoted with non-bold italic font, e.g., $x$. All proofs appear in the appendix.

We start with description of the sensitive data release problem. We then define privacy and utility scores for the problem, which are two possibly competing targets.

\subsection{Problem Description}

\begin{figure}
    \centering
    \includegraphics[width=0.48\textwidth]{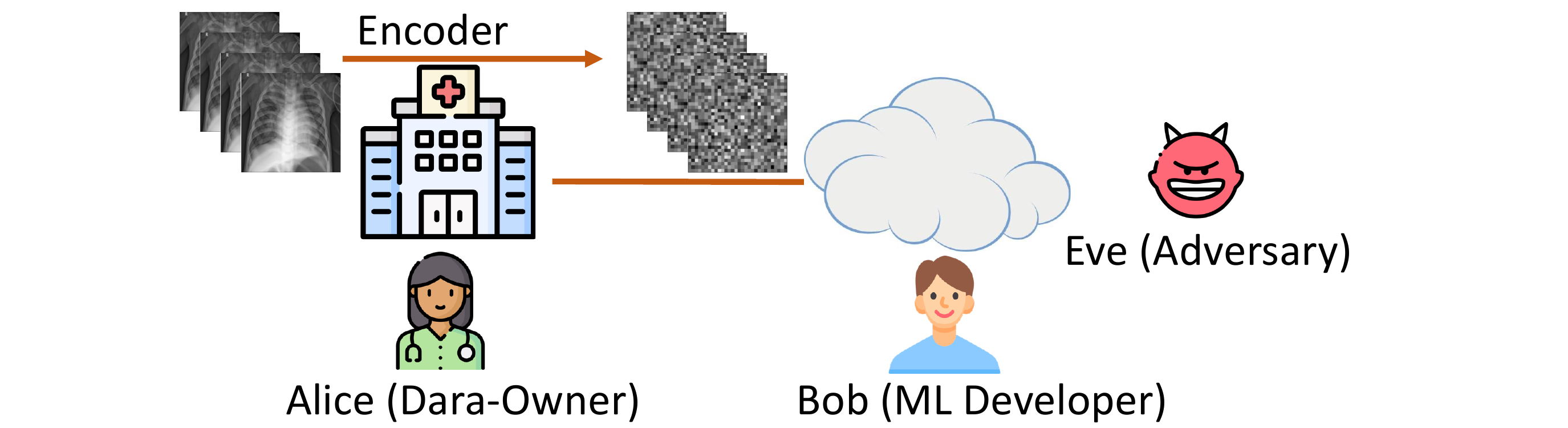}
    \caption{Alice (data-owner) transmits her labeled encoded data to Bob (ML developer). Eve (adversary) attempts to identify information about Alice's raw data beyond their labels.}
    \label{fig:alice-bob-eve}
\end{figure}

We denote the set of all samples by $\mathcal{X}$ and assume it is a finite set. Each sample $x \in \mathcal{X}$ is labeled by a labeling function $L: \mathcal{X} \rightarrow \mathcal{Y}$, where the set of labels $\mathcal{Y}$ is finite. We consider a model with three participants, Alice, Bob and Eve, to be consistent with the common terminology of privacy  (see Fig.~\ref{fig:alice-bob-eve}). We consider a setting where a data-owner (Alice) has some sensitive set of samples $\mathcal{X}_A\subset\mathcal{X}$. Alice needs to train a classifier on her sensitive data to estimate the labeling function $L(\cdot)$. To do so, she wishes to communicate her labeled data $\{(x,L(x))\}_{x\in\mathcal{X}_A}$ with an ML developer (Bob) for learning. However, we assume an adversary (Eve) is also able to observe all communication that take place between Alice and Bob. In general, we consider the information revealed about Alice's data beyond their labels is privacy leakage and must be limited.

\begin{assumption}\label{assum_uniformsamples}
For the ease of theoretical derivations, we assume Alice's samples given their cardinality are chosen uniformly and independently from $\mathcal{X}$. Thus, $\Pr[\bm{\mathcal{X}_A}=\mathcal{X}_A]=\mathbb{C}$, where $\mathbb{C}$ is independent of the realization of $\bm{\mathcal{X}_A}$.
\end{assumption}

Occasionally, we attempt to protect some sensitive features about Alice's data (other than the published labels). For this purpose, let $S: \mathcal{X} \rightarrow \tilde{\mathcal{Y}}$ be another labeling function that describes sensitive features of samples in $\mathcal{X}$, where the set $\tilde{\mathcal{Y}}$ is finite. We assume that Eve has prior knowledge as a public dataset with identical distribution as Alice's data, denoted with $\mathcal{P}\subset\mathcal{X}$. Each public sample is associated with some labels, including $L(x)$ and $S(x)$ for every $x\in\mathcal{P}$.

We consider the following type of schemes: Alice chooses a one-to-one encoding function $T_A:\mathcal{X} \rightarrow \mathcal{Z}$ at random, from a family of functions $\mathcal{F}$ according to distribution $\Pr[\bm{T_A}=T_A]$. Thus, 
$$\mathcal{F}=\{T:\Pr[\bm{T_A}=T]\neq0\}.$$  
She then transmits $\mathcal{O}_{T_A}(\mathcal{X}_A)=\{(T_A(x),L(x))\}_{x\in\mathcal{X}_A}$ to Bob. Bob then trains an ML classifier on the encoded data, thus seeking to learn $L_A = L \circ T^{-1}_A$ on the encoded space $T_A(\mathcal{X})$. In this setting, a scheme is the distribution according to which Alice chooses $T_A$. Eve also receives Alice's labeled encoded data $\mathcal{O}_{T_A}(\mathcal{X}_A)=\{(T_A(x),L(x))\}_{x\in\mathcal{X}_A}$, which we call Eve's observation. We assume Eve also knows the scheme, so she has access to $\Pr[\bm{T_A}=T_A]$, but not exactly the transform sampled by Alice. Thus, we define Eve's prior knowledge $\mathcal{K}_e$ as follows:
\begin{equation*}
       \mathcal{K}_e=\{\{(x,L(x),S(x))\}_{x\in\mathcal{P}},\Pr[\bm{T_A}=T_A]\}.
\end{equation*}

The Shannon entropy $\ent[\bm{x}]$ of a random variable $\bm{x}$ quantifies the average level of uncertainty inherent in its possible outcomes \cite{Shannon48}. By definition, given the distribution of $\bm{x}$ and the space set $\mathcal{X}$ (set of possible values that the random variable $\bm{x}$ can take),
\begin{equation*}
    \ent[\bm{x}]=-\sum_{x\in\mathcal{X}}\Pr[\bm{x}=x]\log \Pr[\bm{x}=x].
\end{equation*}
The conditional entropy $\ent[\bm{x_1}|\bm{x_2}]$ quantifies the average level of \textit{uncertainty} inherent in the possible outcomes of a random variable $\bm{x_1}$ when the outcome of another (possibly correlated) random variable $\bm{x_2}$ is known, and $\ent[\bm{x_1}|\bm{x_2}]=\sum_{x}\ent[\bm{x_1}|\bm{x_2}=x]\Pr[\bm{x_2}=x]$.

\subsection{Privacy Definition and Eve's Attacks}\label{sec:privacy_score}

Given Eve's prior knowledge $\mathcal{K}_e$ and observation $\mathcal{O}_{T_A}(\mathcal{X}_A)$, in this paper, we are interested in what she learns about Alice's private encoder $\bm{T_A}$. Since each $T_A\in\mathcal{F}$ is a one-to-one function from $\mathcal{X}$ to $\mathcal{Z}$, if Eve identifies $T_A$, she can revert back Alice's encoded data into their sensitive original representation. Eve uses her probability distribution on $T_A$ given her prior knowledge and her observations, i.e., 
\begin{equation}
    \label{eq:post_prob}
    P(T)=\Pr[\bm{T_A} = T \mid \mathcal{O}_{T_A}(\mathcal{X}_A),\;\mathcal{K}_e],
\end{equation}
to break Alice's encoding scheme. 

Given Eve's prior knowledge and observations, she chooses an encoder that has the highest likelihood to be Alice's encoder. Therefore the \textit{optimal attack} of Eve is defined as follows:
\begin{definition}[Optimal attack]\label{def:optimal-attack}
The optimal attack outputs an encoder that maximizes $P(T)$:
\begin{equation}
\label{eq:map_est}
    T_A^\text{opt}=\arg\max_{T\in\mathcal{F}} P(T). 
\end{equation}
\end{definition}
\noindent When the solution of (\ref{eq:map_est}) is not unique, Eve would randomly choose one of them. 

The optimal attack is possible for Eve when she has access to the actual probability distribution $P(T)$ in (\ref{eq:post_prob}). The \textit{sub-optimal attack} refers to the case when Eve uses a mismatched distribution $Q(T)$ rather then $P(T)$ to obtain the encoder that has the highest likelihood to be Alice's encoder:
\begin{definition}[Sub-optimal attack]\label{def:sub-optimal-attack}
The sub-optimal attack outputs an encoder that maximizes the mismatched distribution:
\begin{equation*}
    T_A^\text{sub-opt}=\arg\max_{T\in\mathcal{F}} Q(T). 
\end{equation*}
\end{definition}

There are several ways to measure the privacy of an encoding scheme. In this paper, we use Shannon entropy to quantify Eve's average uncertainty about Alice's encoder:

\begin{definition}[Privacy score against optimal attack]\label{def:privacy_score_1}
The privacy score of Alice's scheme is defined as:
\begin{equation*}
\begin{split}
    S_\text{privacy}(\bm{T_A})=\ent[\bm{T_A}\mid \bm{{\mathcal{O}_{T_A}(\mathcal{X}_A)}},\;\mathcal{K}_e].
\end{split}
\end{equation*}
\end{definition}
\noindent A higher privacy score is better as it is equivalent to a higher uncertainty of Eve about Alice's encoder, and consequently a more private encoding scheme. By definition, we have
\begin{equation*}
\begin{split}
    &S_\text{privacy}(\bm{T_A})=\\
    &\hspace{-0.08cm}\sum_{\mathcal{O}_{T_A}(\mathcal{X}_A)}\hspace{-0.15cm}\ent[\bm{T_A}\mid {\mathcal{O}_{T_A}(\mathcal{X}_A)},\;\mathcal{K}_e] \Pr[{\bm{\mathcal{O}_{T_A}(\mathcal{X}_A)}=\mathcal{O}_{T_A}(\mathcal{X}_A)}]{=}\\
    &-\hspace{-0.15cm}\sum_{\mathcal{O}_{T_A}(\mathcal{X}_A)}\sum_{T\in\mathcal{F}}P(T)\log P(T) \Pr[{\bm{\mathcal{O}_{T_A}(\mathcal{X}_A)}=\mathcal{O}_{T_A}(\mathcal{X}_A)}].
\end{split}
\end{equation*}
If the privacy score is zero, there is a unique solution for the optimal attack, which is indeed Alice's encoder. 

Later in Corollary~\ref{cor:matching}, we show that if Alice's encoder can be identified using Alice's original data and encoded data, the privacy score is equivalent to the average uncertainty of Eve about Alice's original data. In general, the privacy score is a lower-bound on the expected number of guesses for an adversary that sorts the encoders from most-likely to least-likely according to (\ref{eq:post_prob}), to correctly guess Alice's encoder \cite{Massey}. The expected number of guesses can be more tightly bounded by Rényi entropy \cite{Arikan,SerdarBOZTAS2014}. As the presented results can be readily extended to other definitions of entropy, in this paper we use the notion of Shannon Entropy.

The privacy score $S_\text{privacy}(\bm{T_A})$ measures the average uncertainty about Alice's encoder when Eve has access to $P(T)$. The average uncertainty about Alice's encoder when Eve uses $Q(T)$ rather than $P(T)$ is obtained using cross-entropy. The cross-entropy $\CE(P,Q)=-\sum_{T\in\mathcal{F}}P(T)\log Q(T)$ is a non-negative metric, lower-bounded by the entropy $\ent[\bm{T_A}\mid {\mathcal{O}_{T_A}(\mathcal{X}_A)},\;\mathcal{K}_e]=\sum_{T\in\mathcal{F}}P(T)\log P(T)$.
\begin{definition}[Privacy score against sub-optimal attack]\label{def:privacy_score_2}
The privacy score of Alice's scheme when Eve has access to a mismatched distribution $Q(T)$ rather than $P(T)$ is:
\begin{equation*}
\begin{split}
    &\tilde{S}_\text{privacy}(\bm{T_A})=\\
    &-\hspace{-0.15cm}\sum_{\mathcal{O}_{T_A}(\mathcal{X}_A)}\sum_{T\in\mathcal{F}}P(T)\log Q(T) \Pr[{\bm{\mathcal{O}_{T_A}(\mathcal{X}_A)}=\mathcal{O}_{T_A}(\mathcal{X}_A)}].
\end{split}
\end{equation*}
\end{definition}

\subsection{Utility Definition}
The utility goal is to increase how much Bob learns about the labeling of Alice's encoded data, i.e., learning $L_A = L \circ T^{-1}_A$ on the encoded space $T_A(\mathcal{X})$. This can be quantified in a variety ways, and we use Shannon entropy to measure Bob's average uncertainty about the labeling function $L_A:\mathcal{Z}\rightarrow\mathcal{Y}$ given the Alice's labeled encoded data.

\begin{definition}[Utility score]
The utility score of Alice's scheme is defined as:
\begin{equation}
    S_\text{utility}(\bm{T_A})=\ent[\bm{L}]-\ent[\bm{L\circ T_A^{-1}} \mid  \bm{\mathcal{O}_{T_A}({\mathcal{X}_A})}].
\end{equation}
\end{definition}
\noindent A higher utility score is better as it is equivalent to a lower uncertainty of Bob about the labeling function that acts on  Alice's encoded data.

In our experiments, to evaluate the utility of an encoding scheme, we compare the prediction utility of two classifiers that are trained with Alice's labeled original data and Alice's labeled encoded data, respectively. We denote the two classifiers with $L^\text{emp}$ and $L^\text{emp}_A$. The classifier $L^\text{emp}$ approximates the function $L$ and is obtained by training an ML model using Alice's un-encoded data $\mathcal{O}({\mathcal{X}_A})=\{(x,L(x))\}_{x\in\mathcal{X}_A}$. The classifier $L^\text{emp}_A$ approximates the function $L_A = L \circ T^{-1}_A$ and is obtained by training a similar ML model using Alice's encoded data $\mathcal{O}_{T_A}({\mathcal{X}_A})=\{(T_A(x),L(x))\}_{x\in\mathcal{X}_A}$. The performance of a classifier is quantified by comparing the output of the original function and the approximated function for some held-out data $\mathcal{Q}\subset\mathcal{X}$ (i.e., generalization accuracy):
$$\{(L^\text{emp}(x),L(x))\}_{x\in\mathcal{Q}},\quad \{(L_A^\text{emp}(T_A(x)),L(x))\}_{x\in\mathcal{Q}}.$$ 
The comparison can be formulated in several ways such as success rate (ratio of pairs with identical values to the total number of pairs) or area under ROC curve (AUC) for two-class classification problems.

\section{Main Theoretical Results}\label{sec:theory}

In this section, we first present some theoretical results to explore the described privacy-utility problem, in Section~\ref{sec:prelim}. In Section~\ref{sec:arch}, we propose architectural lines to improve the privacy score of an encoding scheme. In Section~\ref{sec:adv_attacks}, we study a tractable approximation of Eve's optimal attack, and the privacy score of an encoding scheme against this sub-optimal attack. Finally, in Section~\ref{sec:connection_data_Enc}, we demonstrate how protecting the private encoder, selected by Alice, is connected to protecting her sensitive data.

\subsection{On Privacy of Randomly Encoded Data}\label{sec:prelim}

Here, we identify the set of possible values for Alice's encoder $\Pos[{T_A}]\subseteq\mathcal{F}$, from Eve's point of view. We then use this notion to investigate Eve's probability distribution $P(T)$ and her optimal attack on Alice's encoder. 

\begin{definition}\label{def:Poss_T}
The set of possible values for Alice's encoder given Eve's observations is:
\begin{equation*}
    \begin{split}
        \Pos[{T_A}]&\triangleq \{ T \in \mathcal{F}:  \exists \Bar{X}_T \subset\mathcal{X} \text{ with }\\
        &\{(T(x),L(x))\}_{x\in\Bar{X}_T}=\mathcal{O}_{T_A}({\mathcal{X}_A})\}.
    \end{split}
\end{equation*}
\end{definition}

The distribution of Eve on Alice's encoder given her observations and knowledge can be described in terms of $\Pos[{T_A}]$:
\begin{theorem}\label{theorem:prob_poss}
The probability distribution $P(T)$ can be stated as 
\begin{equation*}
    P(T)=\Pr[\bm{T_A} = T \mid \mathcal{O}_{T_A}({\mathcal{X}_A}),\mathcal{K}_e]=\frac{\Pr[\bm{T_A}=T]}{\Pr[\bm{T_A}\in\Pos[{T_A}]]},
\end{equation*}
if $T\in\Pos[{T_A}]$. $P(T)=0$, otherwise.
\end{theorem}

\begin{corollary}\label{cor:opt_attack}
The optimal attack can also be characterized in terms of $\Pos[T_A]$ as follows:
\begin{equation*}
    T_A^\emph{\text{opt}}=\arg\max_{T\in\Pos[T_A]} \Pr[\bm{T_A} = T].\vspace{-0.17cm}
\end{equation*}
\end{corollary}

\subsection{Architectural Lines to Improve Privacy}\label{sec:arch}

We start from a family of functions $\mathcal{F}$ from which Alice chooses her encoder according to $\Pr[\bm{T_A}=T]$. We wish to understand what kind of operations Alice can perform to improve the privacy of her encoding scheme. To this end, we explore several ways to grow $\mathcal{F}$. In our next proposition, we show that adding arbitrary functions to $\mathcal{F}$ might actually worsen the privacy score.

\begin{proposition} \label{prop:addingcanhurt} 
There exists cases where growing the family of functions that Alice randomly chooses her encoder from lowers the privacy score.
\end{proposition}

However, as we now show, composing families of functions can only preserve or increase the privacy score.

\begin{theorem} \label{theorem:compositiondoesnothusrt}
Let $\mathcal{F}$ and $\mathcal{F}'$ be two families of encoders from which Alice samples her encoder from according to $\Pr[\bm{T_A}=T_A]$ and $\Pr[\bm{T'_A}=T'_A]$, respectively. Consider a new family of encoders $\mathcal{F}''=\mathcal{F}' \circ \mathcal{F} = \{ T' \circ T : T' \in \mathcal{F}', T \in \mathcal{F} \}$, and the associated distribution $$\Pr[\bm{T''_A}=T''_A]=\sum_{T' \circ T=T''_A}\Pr[\bm{T'_A}=T']\Pr[\bm{T_A}=T].$$ 
Then, the privacy score of the new encoding scheme is at least equal to the privacy score of each initial one, i.e.,
\begin{equation*}
\normalfont
    S_\text{privacy}(\bm{T_A})\leq S_\text{privacy}(\bm{T''_A}),\quad S_\text{privacy}(\bm{T'_A})\leq S_\text{privacy}(\bm{T''_A}).
\end{equation*}
\end{theorem}

Theorem \ref{theorem:compositiondoesnothusrt} shows that composing families of encoders cannot reduce the privacy score. Indeed, as we show in the following example, it can potentially increase it.

\begin{example}
Let consider the universe of all samples is $\{(x,L(x))\}_{x\in\mathcal{X}}=\{(1,+),(2,+),(3,-),(4,-)\}$, and consider these two families of encoders, from which Alice uniformly chooses her encoder:

\begin{equation*}
\mathcal{F}=\begin{cases}
T_1:&(1,2,3,4)\\ 
T_2:&(2,1,3,4)
\end{cases},\quad \text{and} \quad
\mathcal{F}'=\begin{cases}
T'_1:&(1,2,3,4)\\ 
T'_2:&(1,2,4,3)
\end{cases}.
\end{equation*}
Now, we consider the composition of these two families of encoders:
\begin{equation*}
    \mathcal{F'}\circ\mathcal{F}=\begin{cases}
        T'_1\circ T_1:&(1,2,3,4)\\ 
        T'_1\circ T_2:&(2,1,3,4)\\ 
        T'_2\circ T_1:&(1,2,4,3)\\ 
        T'_2\circ T_2:&(2,1,4,3)
        \end{cases}
\end{equation*}
We can verify that
\begin{equation*}
    \ent[\bm{T_A}\mid \bm{\mathcal{O}_{T_A}(\mathcal{X}_A)},\mathcal{K}_e]=\ent[\bm{T'_A}\mid \bm{\mathcal{O}_e(T'_A)},\mathcal{K}_e]=1.
\end{equation*}
However, $\ent[\bm{T'_A\circ T_A}|\bm{\mathcal{O}_{T'_A\circ T_A}(\mathcal{X}_A)},\mathcal{K}_e]=2$, which is a higher privacy score than the one for the individual schemes.
\end{example}

We leverage our theoretical results on function composition to guide the design of our encoding scheme. Starting with a weak encoder, a random linear transform, we iteratively enrich the privacy of our scheme through function composition (e.g  by adding with additional non-linear and linear layers), to build a random deep neural network. The non-linearity is a necessary component to happen between the composed layers, otherwise the composed linear layers reduce to just a one-layer linear encoder that is weak. More details on the encoding scheme are presented in Section~\ref{sec:method}.

\subsection{Approximating the Optimal Attack}\label{sec:adv_attacks}

The probability distribution used by the optimal attack is given in (\ref{eq:post_prob}) and is revisited here,
\begin{equation*}
    P(T)=\Pr[\bm{T_A} = T \mid \mathcal{O}_{T_A}(\mathcal{X}_A),\;\mathcal{K}_e].
\end{equation*}

In practice, Eve does not have access to $P(T)$ to perform the optimal attack. However, it has access to public dataset $\mathcal{P}\subset\mathcal{X}$ with samples with the same distribution as Alice's samples. The public samples and Alice's samples have the same distribution even after the transformation via $T_A$. Thus, a tractable solution to approximate the optimal attack, i.e., a sub-optimal attack, is to identify an encoder that results in a similar distribution to distribution of Alice's encoded data when applied to an available public dataset $\mathcal{P}\subset\mathcal{X}$.

There are a variety of ways to quantify the mismatch between two distributions, e.g., Kullback–Leibler (KL) divergence and Kolmogorov-Smirnov distance (K-S) \cite{Kullback59,Karson}. Here, we use the generic form $\dist[\mathcal{Z},\mathcal{Z}']$, where samples of $\mathcal{Z}$ and $\mathcal{Z}'$ are drawn from distributions $p$ and $q$, to represent the mismatch between two distributions $p$ and $q$. Thus, the sub-optimal attack is,
\begin{equation}\label{eq:dist_attack}
        T_A^\text{sub-opt}\triangleq \arg\min_{ T \in \mathcal{F}} \dist[T_A(\mathcal{X}_A),T(\mathcal{P})].
\end{equation}

In fact, the mismatch between distributions of Alice's encoded data $T_A(\mathcal{X}_A)$ and public encoded data via the current estimated encoder, can be used as a loss function for training an ML model that approximates $T_A$. Let assume $T_\theta$ represents all possible choices that can be obtained by the attack model of Eve, where each realization of $\theta$ corresponds to one possible choice for the weights of the model. Thus,
\begin{equation*}
\begin{split}
    &T_A^\text{sub-opt}=T_{\theta^\star},\\
    &\theta^\star=\arg \min_{\theta}  \dist[T_A(\mathcal{X}_A),T_\theta(\mathcal{P})].
\end{split}
\end{equation*}

One popular practical method to measure the distribution mismatch $\dist[T_A(\mathcal{X}_A),T_\theta(\mathcal{P})]$ is the maximum mean discrepancy (MMD) which has recently received significant attention \cite{gretton2012kernel,NIPS2016_5055cbf4}. By definition,
\begin{equation*}
    \text{MMD}[p,q]=\sup_{h\in\mathcal{H}}(E_{x\sim p}[h(x)]-E_{x\sim q}[h(x)]).
\end{equation*}
If $\mathcal{H}$ is the space of bounded continuous functions on $\mathcal{X}$, the MMD measure is zero if and only if $p=q$ \cite{gretton2012kernel}. An estimate of the above measure is obtained by replacing the expectations with empirical average. For having a rich family of functions $\mathcal{H}$, we use the functions in the unit ball of a reproducing kernel Hilbert space (RKHS), as suggested in \cite{4914,10.5555/1756006.1859901}.

The attack (\ref{eq:dist_attack}) is a tractable approximation of optimal attack, and can be considered as if Eve uses another distribution such as $Q$ rather than $P$ for her attack. We remind that the privacy score against the optimal attack $S_\text{privacy}(\bm{T_A})$ and the privacy score against the sub-optimal attack $\tilde{S}_\text{privacy}(\bm{T_A})$ measure the average uncertainty about Alice's encoder when Eve has access to the actual probability distribution $P$ and the approximated distribution $Q$, respectively, see Section~\ref{sec:privacy_score}. In fact,
\begin{equation*}
\begin{split}
    \CE(P,Q){-}&\ent[\bm{T_A}\mid {\mathcal{O}_{T_A}(\mathcal{X}_A)},\;\mathcal{K}_e]{=}\sum_{T\in\mathcal{F}}P(T)\log \frac{P(T)}{Q(T)}\\
    &=D_\text{KL}(P||Q).
\end{split}
\end{equation*}
Here, $D_\text{KL}(P||Q)\geq0$ is the KL divergence. Therefore, 
\begin{equation*}
\begin{split}
&\tilde{S}_\text{privacy}(\bm{T_A})-S_\text{privacy}(\bm{T_A})=\\
&\sum_{\mathcal{O}_{T_A}(\mathcal{X}_A)}\hspace{-0.15cm}D_\text{KL}(P||Q) \Pr[{\bm{\mathcal{O}_{T_A}(\mathcal{X}_A)}=\mathcal{O}_{T_A}(\mathcal{X}_A)}]\geq0.
\end{split}
\end{equation*}
The privacy score of a scheme against sub-optimal attack, who has  $Q(T)$ rather than $P(T)$, is an upper-bound for the actual privacy score. The gap is equivalent to the  KL divergence between the actual distribution and the approximated distribution that Eve uses, averaged over all possible observations. In our experiments, Eve uses the practical MMD measure to close the gap between the two distributions.

\subsection{Protecting the Encoder versus Protecting the Data}\label{sec:connection_data_Enc}

In this subsection, we show how the privacy score, which is the average uncertainty of Eve about Alice's encoder, is connected to her average uncertainty about Alice's original data. We then demonstrate for which encoding schemes these two notions of uncertainty coincide.

\begin{proposition}\label{lemma:alternative_privacy_Score}
The privacy score in Definition~\ref{def:privacy_score_1}, can be written as
\begin{equation*}
\begin{split}
    S_\text{privacy}(\bm{T_A})&=\ent[\bm{\mathcal{X}_A}\mid \bm{{\mathcal{O}_{T_A}(\mathcal{X}_A)}},\;\mathcal{K}_e]\\
    &+\ent[\bm{T_A}\mid \bm{\mathcal{X}_A},\bm{{\mathcal{O}_{T_A}(\mathcal{X}_A)}},\;\mathcal{K}_e].
\end{split}
\end{equation*}
\end{proposition}

\begin{corollary}\label{cor:matching}
    Eve has the same average uncertainty about Alice's encoder and about Alice's sensitive data if
    \begin{equation}\label{cond:matching}
        \ent[\bm{T_A}\mid \bm{\mathcal{X}_A},\bm{{\mathcal{O}_{T_A}(\mathcal{X}_A)}},\;\mathcal{K}_e]=0
    \end{equation}
\end{corollary}
According to Corollary~\ref{cor:matching}, if the realization of encoder can be identified given un-ordered sets of original data and encoded data, the privacy score of the encoding scheme is equivalent to the average uncertainty of Eve about Alice's data. In Section~\ref{sec:neuracrypt}, we provide experimental results showing that (\ref{cond:matching}) holds for the version of \name that will be presented in Section~\ref{sec:method}. We note that (\ref{cond:matching}) does not hold for the encoding schemes that are difficult to break, without protecting the sensitive data. Examples of such useless encoding schemes are those that append random data to each sensitive sample.

Lastly, we note that encoding schemes that satisfy (\ref{cond:matching}) are vulnerable when the adversary has access to un-ordered sets of original and encoded samples, e.g., the cases where $\mathcal{X}_A\cap \mathcal{P}\neq \emptyset$. In fact, having non zero intersection between the public data and private data might lead to zero privacy score for these schemes. Thus, if $\mathcal{X}_A\cap \mathcal{P}\neq \emptyset$, the data-owner either needs to exclude the public samples from her dataset before encoding, to use an encoding scheme with $\ent[\bm{T_A}\mid \bm{\mathcal{X}_A},\bm{{\mathcal{O}_{T_A}(\mathcal{X}_A)}},\;\mathcal{K}_e]>0$ such as \cite{Syfer}, or to use non-invertible transforms as her encoding scheme.

\section{Method: Random DNNs for Image and Text}\label{sec:method}

In this section, we develop two encoding schemes inspired by our result that functional composition can improve the privacy of encoding schemes. The schemes will be presented in two separate subsections: one for encoding image data and one for encoding text data. Later, in Section~\ref{sec:exp}, we provide an empirical evaluation of the privacy and utility performance of the proposed encoding schemes, and will show that our schemes offer improved privacy over linear randomized encoding schemes when:
\begin{equation}\label{assump:const}
    \mathcal{X}_A\cap\mathcal{P}=\emptyset.
\end{equation}

\subsection{Random CNN for Encoding Image Data}

\begin{figure}
    \centering
    \includegraphics[width=0.47\textwidth]{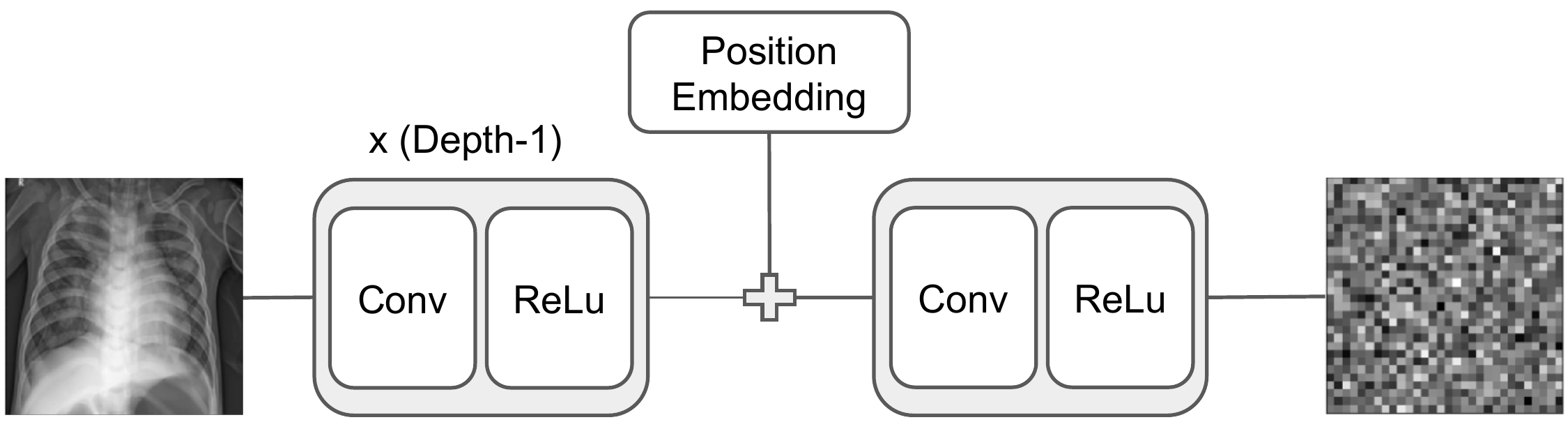}
    \caption{Architecture of image encoder. The encoding task resembles an inference task using an untrained CNN.}
    \label{fig:architecture}
\end{figure}

In this subsection, we focus on imaging tasks, and thus implement our encoders as convolutional neural networks (CNNs). Our encoder architecture is illustrated in Fig.~\ref{fig:architecture}, and consists of convolutional layers with non-overlapping strides, batch normalization \cite{ioffe2015batch}, and ReLu non-linearities. To encode positional information into the feature space while hiding spatial structure, we add a random positional embedding for each patch before the final convolutional and ReLu layers and randomly permute the patches at the output independently for each private sample. This results in an unordered set of patch feature vectors for each image. We note that this architecture is closely inspired by the design of patch-embedding modules in Vision Transformer networks \cite{dosovitskiy2020image,zhou2021deepvit}.

\subsection{Random RNN for Encoding Text Data}
\begin{figure}
    \centering
    \includegraphics[width=0.48\textwidth]{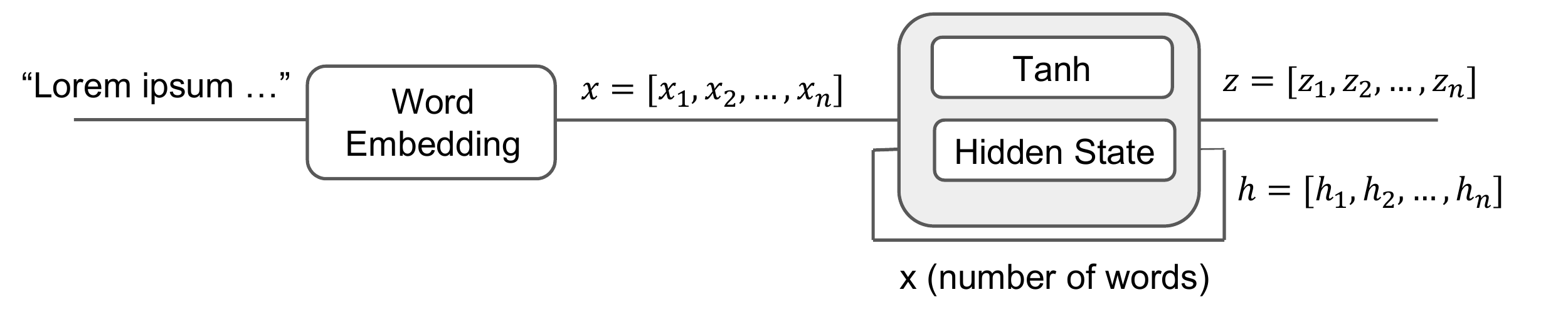}
    \caption{Architecture of text encoder. The encoding task resembles an inference task using an untrained RNN.}
    \label{fig:rnn}
\end{figure}

In this subsection, we focus on natural language processing (NLP) tasks, and thus implement our encoders as random recurrent neural networks (RNNs) \cite{10.1162/neco.1997.9.8.1735}. As illustrated in Fig.~\ref{fig:rnn}, we first map the constituents words of the original sample into vectors using a word embedding. Then, we feed the embedded vectors into an RNN which includes a hidden state and a Tanh non-linearity. The initial state of the RNN is randomly chosen and plays the role of the private key. The encoded output can be either the sequence of RNN outputs, or the final value of its hidden state. The first type of output results in a sequence of encoded words per sample, and along with the labels, can be used to train a downstream model with memory, e.g., another RNN or Long Term Short Memory (LSTM) models. The second type of output results in a single vector which is basically the encoded context, and can be used for training memoryless models such as a dense neural network.

We highlight that using a random RNN for encoding the text data is a benchmark here, and one can use more sophisticated NLP models with randomly initialized states for encoding. Besides, although only the initial value of the hidden state is chosen randomly, the next values of the hidden state depend on the previous states and the incoming words. This results in a sequence of pseudo random values for the hidden state. This is similar to a self-synchronising stream cipher \cite{515609} with the difference that the ciphertext does not need to be decoded to be useful. The unfolded version of a randomly-initialized RNN can be seen as a deep neural network with pseudo-random values for its weights.

\subsection{Private Collaborative Learning}

It is known that a richer training dataset results in better predictive utility of ML models. In fact, the current bottleneck in training competent predictive models in many sensitive applications, such as healthcare, is lack of training data, e.g., \cite{nature_inconvenient_truth,Bak2022} among many others. Sufficient and diverse data for training often does not belong to a single data-owner, and multiple cohorts are encouraged to collaboratively provide data for the training job. 

Let assume we have $D$ data-owners. We wish to enable each data-owner with index $d$, $d\in\{0,\dots,D-1\}$, to publish their labeled dataset $\mathcal{X}_d$ while hiding its sensitive information. Given the dataset $\{(x,L(x))\}_{x\in\mathcal{X}_d}$, the data-owner randomly samples a private encoder $T_d$ according to $\Pr[\bm{T_A}=T_d]$, and uses $T_d$ to produce labeled encoded samples $\{(T_d(x),L(x))\}_{x\in\mathcal{X}_d}$. The data-owner can then deposit the labeled encoded dataset publicly for untrusted third parties for collaborative learning. In this section, we show that multiple data-owners can seamlessly collaborate to develop joint models by publishing datasets on the same task while using independently-sampled encoders, see Fig~\ref{fig:comp}~(c).

From information theoretical view, the average uncertainty of Bob about the labeling function goes down with enriching the training data, i.e., for $d\in\{0,\dots,D-1\}$,
\begin{equation*}
\begin{split}
    \ent[\bm{L} {\mid} \bm{\{(x,L(x))\}_{x\in\mathcal{X}_0}},&\dots,\bm{\{(x,L(x))\}_{x\in\mathcal{X}_{D-1}}}]\\
    &\leq \ent[\bm{L} {\mid}  \bm{\{(x,L(x))\}_{x\in\mathcal{X}_d}}],
\end{split}
\end{equation*}
and in fact, if the combined data is equivalent to the set of all samples, i.e., $\mathcal{X}_0\cup\dots\cup \mathcal{X}_{D-1}=\mathcal{X}$, the entropy of the labeling function given the combined labeled data is zero. 

The same argument holds for training on combined encoded data of multiple data-owners,

\small
\begin{equation*}
\begin{split}
    \ent[\bm{L\circ T_d^{-1}} {\mid} &\cup_{d'} \bm{\bm{\{(T_{d'}(x),L(x))\}_{x\in\mathcal{X}_{d'}}}}]\\
    &\leq \ent[\bm{L\circ T_d^{-1}} {\mid}  \bm{\{(T_d(x),L(x))\}_{x\in\mathcal{X}_d}}],
\end{split}
\end{equation*}
\normalsize

Thus, the utility score obtained by the combined encoded data, is at least equivalent to the utility score when the model is trained with the encoded data of one data-owner. This is because the ML developer can just ignore the data of other data-owners and treat them individually to train $D$ disjoint classifiers. Next, we show as long as knowing the optimal classifier for one data-owner contains useful information for classifying the data of other data-owners, the utility score can theoretically increase by using combined encoded data.

\begin{proposition}\label{prop:multi-data-owner} The utility score of a scheme that uses combined encoded data of multiple data-owners for training is lower bounded by the utility score of the scheme when only encoded data of one data-owner is used for training. The bound is sharp if and only if  
\begin{align*}
    &\ent[\cup_{d'\neq d}\bm{\bm{\{(T_{d'}(x),L(x))\}_{x\in\mathcal{X}_{d'}}}}{\mid} \bm{\{T_d(x)\}_{x\in\mathcal{X}_d},L\circ T_d^{-1}}]\\
    &=\ent[\cup_{d'\neq d}\bm{\bm{\{(T_{d'}(x),L(x))\}_{x\in\mathcal{X}_{d'}}}}{\mid}\bm{\{T_d(x),L(x)\}_{x\in\mathcal{X}_d}}].
\end{align*}
\end{proposition}

Our experiments in Section~\ref{sec:exp} also show improvements in the predictive utility of an ML model when trained with combined individually-encoded data of two data-owners (each one samples its encoder randomly and independently) compared to the case where the model is trained only with encoded data of one data-owner.

\section{Experimental Results}\label{sec:exp}

To evaluate the impact of our randomized encoding scheme on downstream modeling performance, we compared models that are trained with our encoded data to standard architectures trained on raw data. For each classification task and training setting, we report the average AUC of such models. For the case of multiple data-owners, we wished to evaluate the impact of leveraging independently-sampled encoders on modeling accuracy. As a result, we evaluate both model performance when leveraging a single encoder across both data-owners (Combined-Clear), and model performance when leveraging two independent encoders (Combined-Randomized). We note that prediction accuracy in the Combined-Clear setting acts as an upper bound for the Combined-Randomized.

To compare the privacy of multiple encoding schemes, we considered a computational attack that aims to estimate the encoder realization $T_A$. We assumed that the attacker has access to a labeled public dataset $\{(x,L(x)),S(x)\}_{x\in \mathcal{P}}$, and Alice's labeled encoded samples $\{(T_A(x),L(x))\}_{x\in\mathcal{X}_A}$. Given this information, the attacker tries to learn a $T^\text{sub-opt}$ such that $T_A \approx T^\text{sub-opt}$. To estimate $T_A$, we sampled an initial encoder $T_\theta$ with the same architecture as $T_A$, and trained it to minimize the MMD between $Z^*=T_\theta(\mathcal{P})$ (generated ciphertext) and $Z=T_A(\mathcal{X}_A)$ (real ciphertext).

We also consider a scenario where an attacker may try to learn a sensitive attribute classifier on the encoded domain using the estimated encoder and the public data. The attacker can use this classifier on Alice's encoded data $T_A(\mathcal{X}_A)$ to obtain sensitive information that is not released by Alice.

\subsection{Chest X-Ray Data}\label{sec:neuracrypt}
For these experiments, we utilized two benchmark datasets of chest x-rays, MIMIC-CXR (\cite{johnson2019mimic}) and  CheXpert (\cite{irvin2019chexpert}) from Beth Israel Deaconess Medical Center and Stanford, respectively. The MIMIC-CXR and CheXpert datasets are available under the PhysioNet Credentialed Health Data License 1.5.0 license and Stanford University School of Medicine CheXpert Dataset Research Use Agreement, respectively. The samples of each dataset are labeled with up to $5$ abnormalities, i.e., Edema, Pneumothorax, Consolidation, Cardiomegaly and Atelectasis. For each medical condition as a classification task, we excluded exams with an uncertain disease label, i.e., the clinical diagnosis did not explicitly rule out or confirm the disease, and randomly split the remaining data $60{-}20{-}20$ for training, development and testing, respectively. All images were down sampled to $256{\times}256$ pixels. All experiments were repeated $3$ times across different seeds.

\paragraph{Evaluating modeling utility}

For each diagnosis task and training setting, we report the average AUC across the MIMIC-CXR and CheXpert test sets. Our encoding split each image into several patches with size $16{\times}16$, and leveraged a model with depth of $7$ and a hidden dimension of $2048$. This model had $\sim22.9M$ parameters and mapped $256{\times}256$ pixel images to $256{\times}2048$ vectors. Because of the patch-shuffling component of our encoding scheme, the encoded patches are unordered. As a result, we trained Vision Transformers (ViT) \cite{zhou2021deepvit}, a self-attention based architecture that is invariant to patch ordering. Across all experiments, we used a one-layer ViT with a hidden dimension of $2048$ for the utility evaluation, and we compared the classification performance with a raw-sample baseline. We trained all models for $25$ epochs using the Adam optimizer \cite{kingma2014adam}, an initial learning rate of $1\mathrm{e}{-04}$, weight decay of $1\mathrm{e}{-03}$ and a batch size of $128$.

We report our results in predicting various medical diagnoses from chest x-ray datasets in Table \ref{tab:xray-expers}. The models trained on encoded data obtained competitive AUCs to our raw-sample baseline across all training settings. In the multi-hospital setting, we found that trained on encoded data was effectively able to leverage the larger training set to learn an improved classifier, despite using separate encoders for each dataset. Our scheme obtained an average AUC increase of 2 and 3 points compared to training only on the MIMIC-CXR and CheXpert datasets, respectively. Moreover, our scheme demonstrated to achieve equivalent performance in the Combined-Clear and Combined-Randomized settings, showing that multiple institutions do not pay a significant performance cost to collaborate via publishing their randomly-encoded data.

\begin{table}[!htb]
\centering
\caption{Chest x-ray prediction tasks across different training settings. All metrics are average ROC AUCs across the MIMIC-CXR and CheXpert test sets. Guides of abbreviations for medical diagnosis: (E)dema, (P)neumothorax, (Co)nsolidation, (Ca)rdiomegaly and (A)telectasis.}
\label{tab:xray-expers}
\begin{tabular}{ccccccc}
\hline
\textit{Data} & E & P & Co & Ca & A & \textit{Average}\\ 
\hline
\multicolumn{7}{c}{\textbf{Train on MIMIC-CXR}}  \\
\hline
orig & 85 $\pm $ 1  & 69 $\pm $ 3 & 74$ \pm $ 2 & 87 $\pm $ 0 & 83 $\pm $ 1 & \textit{80} \\
\hline
enc & 85 $\pm $ 2 & 72 $\pm $ 1 & 72 $\pm $ 1 & 87 $\pm $ 0 & 83 $\pm $ 1 & \textit{80} \\
\hline
\multicolumn{7}{c}{\textbf{Train on CheXpert}} \\
\hline
orig & 82 $\pm $ 1 & 71 $\pm $ 1 & 72 $\pm $ 3 & 83 $\pm $ 1 & 80 $\pm $ 0 & \textit{77} \\
\hline
enc & 84 $\pm $ 1 & 71 $\pm $ 1 & 75 $\pm $ 2 & 82 $\pm $ 1 & 81 $\pm $ 0 & \textit{79} \\
\hline
\multicolumn{7}{c}{\textbf{Train on Combined-Clear}} \\
\hline
orig & 86 $\pm $ 0 & 77 $\pm $ 1 & 76 $\pm $ 2 & 87 $\pm $ 1 & 85 $\pm $ 0 & \textit{82} \\
\hline
enc & 87 $\pm $ 0 & 76 $\pm $ 3 & 78 $\pm $ 1 & 88 $\pm $ 0 & 85 $\pm $ 1 & \textit{83} \\
\hline
\multicolumn{7}{c}{\textbf{Train on Combined-Randomized}} \\
\hline
enc & 87 $\pm $ 1 & 77$\pm $ 3 & 77 $\pm $ 3 & 86 $\pm $ 1 & 84 $\pm $ 1 & \textit{82}\\
\hline
\end{tabular}
\end{table}

\paragraph{Evaluating modeling privacy}
\label{sec:model_privacy}

We performed the attack on convolutional architectures with a depth of $7$. As a baseline, we also performed the attack when using a simple linear encoder, implemented as single convolutional layer. Across all experiments, we used a hidden dimension of $2048$ and trained $T^\text{sub-opt}$ for $25$ epochs. We performed a grid search over different learning rates and weight decay values for each attack. We recorded the validation MMD, as the loss of the MMD attack, against linear encoding scheme and our encoding scheme. Moreovere, we evaluated the attack by measuring the normalized MSE between generated ($Z^*$) and real ciphertext ($Z$) for some held-out plaintext images. For normalization, we use the MSE of a naive estimation of Alice's encoder that maps every sample to the average of Alice's encoded samples. We report the performance of our adversarial attack in Table~\ref{tab:attack1}. 

\begin{table}[!htb]
\centering
\caption{Evaluation of an adversarial attack on two different encodings for held-out dataset $\mathcal{Q}$.}
\label{tab:attack1}
\begin{tabular}{ c c c}
\hline
Encoding & Validation MMD &  Normalized MSE \\ 
\hline
Linear & 1.139 &  $0.43 \pm 0.01$ \\ 
\hline
Depth-7 & 1.637 & $4.44 \pm 0.12$     \\
\hline
\end{tabular}
\end{table}

To compare the performance of multiple encoding schemes against the second type of attack, we began with the best estimated $T^\text{sub-opt}$ from our adversarial attack experiments, and built a new classifier to predict a sensitive feature of an encoded sample. To build the training data, we use a public data and $T^\text{sub-opt}$ to obtain $Z^*$. We report the ROC AUC of this classifier on both $Z$ and $Z^*$.  We performed this attack for both an encoding with a depth of $7$ and a hidden dimension of $2048$, as leveraged in the modeling utility experiments, and when using a linear encoding. For each experiment, we trained a ViT for $25$ epochs using the Adam optimizer, an initial learning rate of $1\mathrm{e}{-04}$ and a batch size of $128$. We report the performance of our sensitive feature attack in Table~\ref{tab:attack2}. As expected, using a linear encoding is not robust to both attacks.

\begin{table}[!htb]
\centering
\caption{Performance of sensitive feature attack on Linear and our encodings.}
\label{tab:attack2}
\begin{tabular}{ccc}
\hline
Encoding & AUC on $Z^*$ & AUC on $Z$\\ 
\hline
Linear & 89 $\pm$ 1 & 86 $\pm$ 1 \\ 
\hline
Depth-7 & 84 $\pm$ 1 & 52 $\pm$ 4 \\ 
\hline
\end{tabular}
\end{table}

\paragraph{Verification of Equality (\ref{cond:matching})}

Here, we empirically validate that for the described encoding scheme, we have
$\ent[\bm{T_A}\mid \bm{\mathcal{X}_A},\bm{{\mathcal{O}_{T_A}(\mathcal{X}_A)}},\mathcal{K}_e]\approx 0,$ 
and thus Alice's data and Alice's encoder have the same average uncertainty for the attacker. For this purpose, we show that an attacker who has access to $\mathcal{X}_A$, $\mathcal{O}_{T_A}(\mathcal{X}_A)$, and $\mathcal{K}_e$ can recover $T_A$. We realize this goal in two steps. Using the un-ordered datasets $\mathcal{X}_A$ and $\mathcal{O}_{T_A}(\mathcal{X}_A)$, we first demonstrate that an attacker can use her knowledge $\mathcal{K}_e$ to recover the pairs $\{(x, T_A(x))\}_{x \in \mathcal{X}_A}$. Then, using a sufficient number of matching pairs $\{(x, T_A(x))\}_{x \in \mathcal{X}_A}$, we recover $T_A$.

For the first step, we leverage a model-based attacker $M$, trained using the following procedure: 
\begin{enumerate}
\item Sample an encoder $T\in\mathcal{F}$ according to $\Pr[\bm{T_A}=T]$. 
\item Create an encoded dataset $T(\mathcal{X}_A)$. 
\item Update the weights of the model $M$ such that for every $x,y\in\mathcal{X}_A$, where $x\neq y$,  $M(x,T(x))$ is high (high similarity), while
$M(y,T(x))$ is low (low similarity).
\item Repeat from 1) until convergence.
\end{enumerate}
We emphasize that at each iteration of the algorithm, a new encoding function $T$ is sampled. Hence, $M$ learns to generalize across the family of encoding functions for the fixed dataset $\mathcal{X}_A$. As a result, $M$ also generalizes to Alice's specific encoder $T_A$ and the attacker obtains pairs of plaintexts and their corresponding ciphertexts.

For the second step, we utilize the corresponding pairs to carry a plaintext attack using gradient descent on the weights of a candidate encoder $T$, to minimize the loss between $T(x)$ and $T_A(x)$ for every  $x\in\mathcal{X}_A$.

For an encoder architecture with depth $7$, in the first step (the matching game), we observed the re-identification AUC $99.93$ and re-identification accuracy $89.44$, respectively. As for the second step, we measured the success of recovering Alice's encoder by mean square error (MSE) metric. We observed a validation MSE of 0.0435 between pairs of samples in $\{T_A(x),T(x)\}_{x\in\mathcal{X}_A}$. In comparison, the MSE between $\{T_A(x),T_R(x)\}_{x\in\mathcal{X}_A}$, where $T_R$ is a random encoder with the same architecture as $T_A$ is 0.780.

\subsection{SMS Text Data}
The dataset contains labeled English SMS messages for spam classification \cite{spamdata}. A word-level tokenizer was used to tokenize the samples (messages). Then, samples with less than $5$ tokens were removed, in order to train only on texts with discernible meaning. The filtered dataset is composed of $4916$ samples. Next, we used GloVe~$200$ disctionary as the word embedding \cite{pennington2014glove}. GloVe~$200$ is trained on Wikipedia and GigaWord~$5$ data to transform words into a vector representations, such that some similarity metrics between the words are preserved. We simulate two data-owners by equally splitting this dataset into two - Spam~Dataset~1 and Spam~Dataset~2. We randomly split each dataset $45{-}45{-}10$ for training, development and testing, respectively. Our encoder is a randomly initialized RNN model with $200$ features in its hidden state. The final value of the hidden state represents the encoded sample we use for training a classifier as described next. 

\paragraph{Evaluating modeling utility}
The resulting encoded representations associated with their labels are sent to a downstream densely connected neural network for spam classification. We evaluate the utility of our encoder by computing the AUC of spam classification trained on encoded data, compared to a raw-sample baseline that is trained on original uncoded data. We train the models using early stopping with a patience of $10$ points to avoid overfitting, and we use the Adam optimizer and learning rate of $1\mathrm{e}{-03}$. 

\begin{table}[!htb]
\centering
\caption{Spam prediction tasks across different training settings. All metrics are average ROC AUCs across SMS Spam test sets.}
\label{tab:spam-utility}
\begin{tabular}{cc}
\hline
\textit{Data} & AUC\\ 
\hline
\multicolumn{2}{c}{\textbf{Train on Spam Dataset 1}}  \\
\hline
orig & 99.48 \\
\hline
enc & 88.74 \\
\hline
\multicolumn{2}{c}{\textbf{Train on Spam Dataset 2}} \\
\hline
orig & 97.96 \\
\hline
enc & 89.55 \\
\hline
\multicolumn{2}{c}{\textbf{Train on Combined-Clear}} \\
\hline
orig & 99.37 \\
\hline
enc & 89.55 \\
\hline
\multicolumn{2}{c}{\textbf{Train on Combined-Randomized}} \\
\hline
enc & 87.39\\
\hline
\end{tabular}
\end{table}

\paragraph{Evaluating modeling privacy}
To test the privacy of the encoder, we conduct an adversarial MMD attack to obtain an estimated encoder, and recorded the validation MMD as well as AUC of obtaining sensitive feature from Alice's encoded data using the estimated encoder. For the sensitive feature attack, we used the same label (spam or not) as both public label and private label since we did not have access to a text dataset with two features suitable for our experiments. We trained a classifier on public data encoded via the estimated encoder, and used the trained classifier to estimate the label of Alice's encoded data. The results are given in Table~\ref{tab:attack3} and show that linear encoding is notably weaker than our RNN-based encoding, against both attacks.

\begin{table}[!htb]
\centering
\caption{Adversarial attacks on two different encodings. Left: MMD Between $T_A(\mathcal{Q})$ and $T^\text{sub-opt}(\mathcal{Q})$ for held-out dataset $\mathcal{Q}$. Right: Performance of sensitive feature attack.}
\label{tab:attack3}
\begin{tabular}{ c c}
\hline
Encoding &  Validation MMD \\ 
\hline
Linear &  0.0472 \\ 
\hline
R-RNN  & 0.0512 \\ 
\hline
\end{tabular}\hspace{1cm}
\begin{tabular}{cc}
\hline
Encoding & AUC\\ 
\hline
Linear & 66.14 \\ 
\hline
R-RNN & 52.98 \\ 
\hline
\end{tabular}
\end{table}

\section{Discussion}\label{sec:Discussion}

Since the leading results of using a single (potentially randomized) transform to encode samples of an ML training dataset, also known as instance encoding, there were a couple of works that support or challenge its privacy promises. We conclude our paper by briefly going through a subset of these works, and argue how \name stands in line of these theoretical and experimental arguments. 

From the information theoretical view, a scheme is perfectly private, if the transformed and original datasets have zero mutual information. However, perfect privacy is not helpful for data sharing, since the encoded data cannot be used for training a classifier which is the authorized use. Recently, in \cite{RassouliTIFS2020,RavivISIT2022}, the notion of perfect sample privacy and perfect subset privacy were considered, which are shown to be attainable using instance encoding. These papers demonstrated the possibility of ensuring zero mutual information between the encoded samples and any subset of original samples with a constrained cardinality, while preserving the learnibility of the encoded dataset. In this paper, we took an alternative approach and changed the focus of privacy question from the original data to the random choice of the encoder taken by the data-owner (i.e., the private key). Thus, the privacy was evaluated against an adversary who is interested to break the encoder given the sensitive encoded data and some relevant uncoded public data.

The arguments that discourage using instance randomized encoding schemes for privacy purposes either targeted a specific encoding scheme, such as linear schemes, or are based on assumptions that do not apply to the setting of this paper. After the introduction of Instahide \cite{instahide}, where the sensitive samples are randomly and linearly mixed together and with some public samples before training, it was shown experimentally in \cite{carlini2020attack} that this random mixing can be un-done. Further, it was argued in \cite{carlini2021private} that it is theoretically impossible to achieve a form of privacy (called distinguishing privacy) by means of mix-up type encodings. This impossibility argument does not apply to \name, as it is neither a linear encoding nor a mix-up type scheme. The results in \cite{carlini2021private} do not consider encoding schemes that use private keys, which are focus of this paper. 

Later in \cite{https://doi.org/10.48550/arxiv.2108.07256}, an attack was proposed to re-identify correspondences of transformed data (encoded via the scheme presented in Section~\ref{sec:neuracrypt}) and raw data, from shuffled datasets \cite{yala_challenge}. The attacker in this task has access to matching (un-ordered) original and encoded datasets, and intends to break the encoder. This is explicitly not the setting we considered for the encoding schemes we presented in Section~\ref{sec:method}, where there is no data overlap. In fact, we proved in Proposition~\ref{lemma:alternative_privacy_Score} that if such an assumption holds, the key and the original data have the same average uncertainty in eyes of an adversary. 

The theoretical impossibility result of \cite{https://doi.org/10.48550/arxiv.2108.07256} which argues the impossibility of achieving ideal privacy for non-trivial encoding schemes is not valid for our setting for two reasons: (i) It is assumed in \cite{https://doi.org/10.48550/arxiv.2108.07256} that the encoding scheme has some auxiliary information about the labeling function, which does not apply to our setting where the encoder is drawn from a distribution that is independent of the data distribution. (ii) We do not target the perfect or ideal privacy in this paper. Instead, we quantify the privacy obtained via a key distribution, using the average uncertainty of the adversary about the encoder chosen by the data-owner given observed encoded data and some public un-encoded data.

Lastly, we note that this paper does not claim proposing an encoding scheme that guarantees the encoded data cannot be used for an un-authorized task (i.e., anything beyond the designated ML training task). Rather, it sheds light on this important privacy-utility problem. In particular, we theoretically showed why the random non-linear transforms, and especially the random deep neural networks, are a viable candidate for this problem. We provided empirical experiments to test the robustness of instances of our encoding scheme against adversarial attacks, following standard practice in security ~\cite{standard2001announcing, dworkin2015sha}. While our empirical attacks are not sufficient to guarantee any privacy, they are insightful to demonstrate trade-offs and measure improvements in the scheme design. Our results validated our presented theoretical argument that composing random functions can offer improved privacy over linear approaches, while maintaining competitive accuracy to raw-sample baselines. Moreover, we demonstrated that multiple independent random encoders can be combined to achieve improved overall utility. While, \name should not be directly applied for private data sharing today, it's promising empirical properties show that randomized neural networks warrant further study, as they offer a new direction of private encoding scheme design.

\section{Conclusions and Future Work}\label{sec:conclusion}

This paper combined the idea of key-based encoding and training on the latent representation of data. The encoded data does not need to be decoded to be used for the downstream training task. Thus, effectively the key does not need to be shared among parties. With the introduced notions of privacy and utility scores of a random encoding scheme, we proposed how one can improve the randomized scheme. While the proposed approach does not guarantee zero information leakage beyond the labeling data, our analysis provided a useful machinery for the design of competent encoding schemes.

On two benchmark chest x-ray datasets, MIMIC-CXR and CheXpert, and a spam text dataset, we found that models trained on our encoded data obtained competitive performance to our raw-sample baselines. In the multi-institutional setting, where each site leverages an independently chosen encoder, we demonstrated that models trained on combined data of multiple cohorts could effectively leverage the larger training data to learn improved classifiers. 

Developing randomized coding schemes which facilitate collaborative training among data-owners with different types of data is a promising future direction to pursue. Another future direction is to develop hybrid collaborative schemes, which allow data-owners to share either their raw data, randomly encoded data, or their model updates. Last but not least, investigating other information measures for quantifying the privacy and utility performance of an encoding scheme remains for the future research.

\bibliographystyle{IEEEtran}
\bibliography{references}

\appendix

This section belongs to the proofs of the theoretical results in this paper.\\

\noindent\textbf{Proof of Theorem~\ref{theorem:prob_poss}.}\\
We first note that according to Definition~\ref{def:Poss_T}, for any $T\notin \Pos[{T_A}]$, we have $\Pr[\bm{T_A} = T \mid \mathcal{O}_{T_A}({\mathcal{X}_A}),\;\mathcal{K}_e]=0$. Next, we consider the case where $T\in \Pos[{T_A}]$:
\small
\begin{align}
    &\Pr[\bm{T_A} = T \mid \mathcal{O}_{T_A}({\mathcal{X}_A}),\mathcal{K}_e]\nonumber\\
    &{=}\frac{\Pr[\bm{\mathcal{O}_{T_A}({\mathcal{X}_A})}=\mathcal{O}_{T_A}({\mathcal{X}_A})\mid \bm{T_A} = T,\mathcal{K}_e]\Pr[\bm{T_A}=T]}{\hspace{0.2cm}\sum_{T'}\Pr[\bm{\mathcal{O}_{T_A}({\mathcal{X}_A})}{=}\mathcal{O}_{T_A}({\mathcal{X}_A}){\mid}\bm{T_A}=T',\mathcal{K}_e]\Pr[\bm{T_A}{=}T']}\label{eq:bayes_lemma_pos}\\
    &=\frac{\Pr[\bm{\mathcal{X}_A}=\Bar{X}_T]}{\hspace{0.2cm}\sum_{T'\in\Pos[T_A]}\Pr[\bm{\mathcal{X}_A}=\Bar{X}_{T'}]\Pr[\bm{T_A}=T']}\Pr[\bm{T_A}=T]\label{eq:existence_lemma_pos}\\
    &= \frac{1}{\sum_{T'\in\Pos[T_A]}\Pr[\bm{T_A}=T']}\Pr[\bm{T_A} = T]\label{eq:equi_probable_lemma_pos}.
\end{align}
\normalsize
Equation (\ref{eq:bayes_lemma_pos}) is the application of the Bayes' theorem, and that by construction $\Pr[\bm{T_A}=T\mid\mathcal{K}_e]=\Pr[\bm{T_A}=T]$. For (\ref{eq:existence_lemma_pos}), we use the fact that each transform in the family is a one-to-one function. Thus given the encoder, the probability of observing a set of encoded data is equivalent to the probability that the equivalent uncoded data be Alice's data. Finally, equation (\ref{eq:equi_probable_lemma_pos}) is the result of the distribution of Alice's data described in Assumption~\ref{assum_uniformsamples}, and the proof is concluded. \qed \\

\noindent\textbf{Proof of Proposition~\ref{prop:addingcanhurt}.}\\
Let $\{(x,L(x))\}_{x\in\mathcal{X}}=\{(1,+),(2,+),(3,-),(4,-)\}$, and consider these two families of encoders, from which Alice uniformly chooses her encoder:
\begin{equation*}
\mathcal{F}=\begin{cases}
T_1:&(1,2,3,4)\\ 
T_2:&(2,1,3,4)\\
T_3:&(1,2,4,3)\\ 
T_4:&(2,1,4,3)
\end{cases},\quad \text{and} \quad
\mathcal{F}'=\begin{cases}
T_1:&(1,2,3,4)\\ 
T_2:&(2,1,3,4)\\
T_3:&(1,2,4,3)\\ 
T_4:&(2,1,4,3)\\ 
T_5:&(3,4,1,2)
\end{cases}
\end{equation*}
We denote with $\bm{T_A}$ and $\bm{T'_A}$ as two random variables representing Alice's encoder, when uniformly distributed over $\mathcal{F}$ and $\mathcal{F}'$, respectively. Here, $\ent[\bm{T_A}\mid {\mathcal{O}_{T_A}(\mathcal{X}_A)},\mathcal{K}_e]=2$ regardless of Alice's data or Eve's observation, and thus $\ent[\bm{T_A}\mid \bm{\mathcal{O}_{T_A}(\mathcal{X}_A)},\mathcal{K}_e]=2$. However, $\ent[\bm{T'_A}\mid \mathcal{O}_{T_A}(\mathcal{X}_A),\mathcal{K}_e]=0$ if for example $(3,+)\in\mathcal{O}_{T_A}(\mathcal{X}_A)$. In general, when Alice's data is non-empty,
\begin{equation*}
    \ent[\bm{T'_A}\mid\bm{\mathcal{O}_{T_A}(\mathcal{X}_A)},\mathcal{K}_e]
    =2\times 4/5 + 0\times 1/5=8/5.    
\end{equation*} \text{ }\qed \\

\noindent\textbf{Proof of Theorem~\ref{theorem:compositiondoesnothusrt}.}\\
We show that if $T'_A$ is given to Eve, then the scheme is as private as if Alice had sampled her encoder from $\mathcal{F}$: 

\begin{equation}
    \begin{split}
        \Pr&[\bm{T'_A \circ T_A}= T \mid \mathcal{O}_{T_A}(\mathcal{X}_A),\;\mathcal{K}_e,\;T_A']\\
        &=\Pr[\bm{T_A} = (T'_A)^{-1}\circ T\mid \mathcal{O}_{T_A}(\mathcal{X}_A),\;\mathcal{K}_e]\\
        &=\Pr[\bm{T_A} = T_A\mid \mathcal{O}_{T_A}(\mathcal{X}_A),\;\mathcal{K}_e].
    \end{split}
\end{equation}
Therefore, we also have
\begin{equation*}
    \begin{split}
        \ent&[\bm{T'_A \circ T_A}\mid \mathcal{O}_{T_A}(\mathcal{X}_A),\;\mathcal{K}_e,\;T_A']=\ent[\bm{T_A}\mid \mathcal{O}_{T_A}(\mathcal{X}_A),\;\mathcal{K}_e].
    \end{split}
\end{equation*}
The above essentially shows the ambiguity about the encoder selected by Alice from $\mathcal{F}' \circ \mathcal{F}$ is at least equal to the ambiguity when Alice selects her encoder from $\mathcal{F}$. Thus, $\mathcal{F}' \circ \mathcal{F}$ has a privacy score that is equal or grater than the privacy score of $\mathcal{F}$. Analogous arguments show that $\mathcal{F}' \circ \mathcal{F}$ has a privacy score that is equal or grater than the privacy score of $\mathcal{F}'$. \qed \\

\noindent\textbf{Proof of Proposition~\ref{lemma:alternative_privacy_Score}}\\
\noindent {By construction, the encoders that Alice can choose from are one-to-one mappings, thus
\begin{equation}\label{eq:results_of_assumptions}
    \begin{split}
        &\ent[\bm{\mathcal{X}_A}\mid \bm{T_A},\bm{{\mathcal{O}_{T_A}(\mathcal{X}_A)}},\;\mathcal{K}_e]=0.
    \end{split}
\end{equation}
Besides, because of the chain rule of the information entropy,
\begin{equation}\label{eq:chain_rule}
    \begin{split}
        &\ent[\bm{\mathcal{X}_A},\bm{T_A}\mid \bm{{\mathcal{O}_{T_A}(\mathcal{X}_A)}},\;\mathcal{K}_e]\\
        &=\ent[\bm{\mathcal{X}_A}\mid \bm{{\mathcal{O}_{T_A}(\mathcal{X}_A)}},\;\mathcal{K}_e]{+}\ent[\bm{T_A}\mid \bm{\mathcal{X}_A},\bm{{\mathcal{O}_{T_A}(\mathcal{X}_A)}},\;\mathcal{K}_e]\\
        &=\ent[\bm{T_A}\mid \bm{{\mathcal{O}_{T_A}(\mathcal{X}_A)}},\;\mathcal{K}_e]{+}\ent[\bm{\mathcal{X}_A}\mid \bm{T_A},\bm{{\mathcal{O}_{T_A}(\mathcal{X}_A)}},\;\mathcal{K}_e]
    \end{split}
\end{equation}
Using (\ref{eq:results_of_assumptions}) and (\ref{eq:chain_rule}), we can conclude the proof.}\qed \\

\noindent\textbf{Proof of Proposition~\ref{prop:multi-data-owner}.}\text{ }\\
\begin{align*}
    &\ent[\bm{L\circ T_d^{-1}}{\mid}  \cup_{d'} \bm{\{(T_{d'}(x),L(x))\}_{x\in\mathcal{X}_{d'}}}]\\
    &{=}\ent[\bm{L\circ T_d^{-1}}{\mid}  \bm{\{(T_{d}(x),L(x))\}_{x\in\mathcal{X}_{d}}}]\\
    &{+}\ent[\cup_{d'{\neq} d}\bm{\{(T_{d'}(x){,}L(x))\}_{x\in\mathcal{X}_{d'}}}{\mid} \bm{T_d(\mathcal{X}_d),L{\circ}T_d^{-1}}]\\
    &{-}\ent[\cup_{d'{\neq} d}\bm{\{(T_{d'}(x){,}L(x))\}_{x\in\mathcal{X}_{d'}}}{\mid} \bm{\{T_d(x),L(x)\}_{x\in\mathcal{X}_d}}],
\end{align*}

\noindent where we used Bayes' theorem for information entropy and the fact that having the function $L\circ T_d^{-1}$ and $T_d(\mathcal{X}_d)$ subsumes the information in $\{(T_d(x),L(x))\}_{x\in\mathcal{X}_d}$. \qed

\end{document}